\documentclass{article}

\usepackage{arxiv}

\usepackage[utf8]{inputenc} % allow utf-8 input
\usepackage[T1]{fontenc}    % use 8-bit T1 fonts
\usepackage{hyperref}       % hyperlinks
\usepackage{url}            % simple URL typesetting
\usepackage{booktabs}       % professional-quality tables
\usepackage{amsfonts,amsmath}       % blackboard math symbols

\usepackage{nicefrac}       % compact symbols for 1/2, etc.
\usepackage{microtype}      % microtypography
\usepackage{lipsum}		% Can be removed after putting your text content
\usepackage{graphicx}
\usepackage[sort&compress,numbers]{natbib}
\usepackage{doi}
\usepackage{caption}
\usepackage{subcaption}
\usepackage{float}
\usepackage{xcolor}
\usepackage{threeparttablex}
\usepackage{marvosym}
\usepackage{algorithm}
\usepackage{algpseudocode}
\newcommand\mymapsto{\mathrel{\ooalign{$\rightarrow$\cr%
  \kern-.15ex\raise.275ex\hbox{\scalebox{1}[0.522]{$\mid$}}\cr}}}

\title{Generative design of multimodal Soft Pneumatic Actuators}

\author{ \href{https://orcid.org/0000-0002-1562-3687}{\includegraphics[scale=0.06]{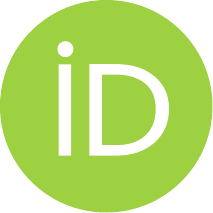}\hspace{1mm}Saswath Ghosh} \\
	Department of Applied Mechanics\\
	Indian Institute of Technology Delhi, \\ New Delhi, India, 110016. \\
	\texttt{S\_Ghosh@am.iitd.ac.in} \\
	\And
	\href{https://orcid.org/0000-0003-2720-4620}{\includegraphics[scale=0.06]{orcid.pdf}\hspace{1mm}Sitikantha Roy}\thanks{Corresponding author} \\
	Department of Applied Mechanics\\
	Indian Institute of Technology Delhi, \\ New Delhi, India, 110016. \\
	\texttt{sroy@am.iitd.ac.in} \\
}

\date{}

\renewcommand{\shorttitle}{PaGD}

\begin{document}
\maketitle

\begin{abstract}
The recent advancements in machine learning techniques have steered us towards the data-driven design of products. Motivated by this objective, the present study proposes an automated design methodology that employs data-driven methods to generate new designs of soft actuators. One of the bottlenecks in the data-driven automated design process is having publicly available data to train the model. Due to its unavailability, a synthetic data set of soft pneumatic network (Pneu-net) actuators has been created. The parametric design data set for the training of the generative model is created using data augmentation. Next, the Gaussian mixture model has been applied to generate novel parametric designs of Pneu-net actuators. The distance-based metric defines the novelty and diversity of the generated designs. In addition, it is noteworthy that the model has the potential to generate a multimodal Pneu-net actuator that could perform in-plane bending and out-of-plane twisting. Later, the novel design is passed through finite element analysis to evaluate the quality of the generated design. Moreover, the trajectory of each category of Pneu-net actuators evaluates the performance of the generated Pneu-net actuators and emphasizes the necessity of multimodal actuation. The proposed model could accelerate the design of new soft robots by selecting a soft actuator from the developed novel pool of soft actuators.
\end{abstract}

% keywords can be removed
\keywords{Design automation \and Soft Actuators \and Generative design \and Multimodal actuation \and Soft robotics}

\section{Introduction}
% introduction include optimization based pneu-net actuators. check once reference correctness\\
% Our method is general...not restricted to bending actuation. Also method can be generalised for other actuators.\\
% Discuss programmable design, automated design papers.\\
% Show results in context of soft robotics instead of ML tools.\\
% gripper would be best possible application of pneu-net. Show some study related to that. Instead, in my understanding, the feasibility of generative design should be based on tasks rather than discrete performance metrics.\\
% Talk about In the context of DDM, the algorithm learns the distribution of data collected from various sources in some multidimensional design manifolds and generates novel designs within the same space which are still nonexistent.  \\
% Connect it with soft robot design. next para write why not optimization?\\
% In the context of DDM, the algorithm learns the distribution of data collected from various sources in some multidimensional design manifolds and generates novel designs within the same space which are still nonexistent.  \\
% Connect it with soft robot design. next para write why not optimization?\\

Soft robots have intrigued researchers in the last decade owing to their potential applications in diverse fields, including medical robotics, exoskeletons, and space exploration. The extensive development of these soft robots that are capable of growing, grasping, manipulating, augmenting human motions, crawling, and jumping in unknown terrains has opened up new possibilities for industries such as healthcare, manufacturing, and exploration \cite{gonzalez2023soft,sambhav2022integrated,boyraz2018overview,zhang2023progress}. As the field advances, a primary focus is on designing and developing soft, flexible actuators to emulate these movements, capable of generating large deformations and forces with minimal input. In that context, the design of soft pneumatic actuators has evolved \cite{xavier2022soft} following the pioneering work of the Whitesides group \cite{mosadegh2014pneumatic} on Pneu-net actuators. Moreover, the design of other soft pneumatic actuators like FREE \cite{singh2020designing,connolly2015mechanical}, textile pneumatic \cite{simpson2017exomuscle}, electro-pneumatic actuators \cite{kumar2021modeling,liu2023optimization} has emerged, presenting intriguing features and showcasing diverse modes of actuation. The design of these soft actuators could be generated by altering their geometrical, material, and structural properties. For example, the geometrical features in the design of Pneu-net actuator could be modified to obtain multimodal actuation  \cite{wang2018programmable}. However, the process of designing a new actuator is not straightforward, relying on human effort and intuition and often consuming a substantial amount of time. Henceforth, exploring the automated design methodology for designing new actuators in tandem with humans is advisable.

Recently, generative models like Generative Adversarial Networks (GAN), Variational Autoencoders (VAE), and transformers have been extensively applied to image and speech synthesis, and Natural Language Processing (NLP) problems to generate images and text \cite{brown2020language}. Especially in the field of NLP, these models have shown potential use cases and developments of chatbots like Chat-GPT and DALL-E. In addition, generative models have been applied to create novel engineering designs, for example, bicycle, aerofoil, wheel, compressor design, and other engineering applications \cite{regenwetter2022biked, yoo2021integrating, shu20203d,rani2024generative}. In line with that, the usage of generative models in robotics design is a way forward towards human-machine collaborative design approach, as described in \cite{stella2023can}. Generative Design (GD) method learns the distribution of design space using existing data and can generate novel designs in contrast to the traditional approach, where the new design relies on human learning and intuition. The significance of GD in various engineering design domains has been studied but its applicability in the context of soft actuator design has yet to be firmly established. Recently, the automated design of soft actuators has been addressed in few works \cite{connolly2017automatic,ellis2022generative,smith2022automated}, by employing an optimization techniques to complete a task. For instance, an automated design of fiber-reinforced soft actuators was used to develop a soft gloves mimicking human hand motion. An optimization algorithm predicts the design of actuators required to move the finger in a trajectory \cite{connolly2017automatic}. Similarly, an optimization technique coupled with finite element solver was employed to generate new designs of soft pneumatic actuators that can provide planar motions \cite{ellis2022generative}. Next, an automated design of bending Pneu-net actuator has been proposed to maximize blocked force and bending angle \cite{smith2022automated}. The article highlights the need towards an automatic design of Pneu-net actuators and proposes a multi-objective optimization technique to achieve so. However, no generative model was implemented that could learn the distribution of design parameters. In similar sense, an optimal design of a soft pneumatic gripper to grasp an object was designed using topology optimization technique \cite{zhang2017design}. Moreover, a structural optimization technique was used to maximize the bending angle and stiffness of soft pneumatic bending  actuators subjected to design-dependent pressure loads \cite{chen2019optimal}. Similarly, an optimal design of soft pneumatic bellow actuators for a particular shape matching could be attained as described in \cite{yao2023design}. Further, a multimodal actuation could be obtained with soft pneumatic actuators by changing its design parameters \cite{wang2018programmable,jiang2021modeling}, thus advancing the domain knowledge. However, all the above studies uses an optimization technique that do not learn the design distribution but rather yield an optimal output for a specific task. Henceforth, the necessity of generative models in designing soft actuators is evident.

As a first step, publicly available large data is required to train the generative design model. Large amounts of publicly available data on different engineering components still need to be made available. Although a few real design data, such as LINKS \cite{heyrani2022links} and BIKED \cite{regenwetter2022biked}, have been published, the availability of different design component data remains limited. Consequently, generating synthetic data through random sampling, data augmentation, or data imputation has emerged as a potential solution to this challenge. In the context of soft actuator design, the challenge remains the same. Moreover, the open-access web toolkit containing data on different soft actuators performance is not accessible \cite{henderson2021data}. To the best of author's knowledge, any design data of either pneumatic or any other soft actuators is not made publicly available. Moreover, the generated design may exhibit similarities to existing designs. The similarity metrics would be helpful to identify the difference between the generated and existing design, i.e., the novelty of the generated design \cite{regenwetter2023beyond}. The design's novelty could be found by calculating the distance of the generated sample from the nearest training data—the higher value results in more novel designs. In addition, the generated design data should be uniform and well spread. The diversity in parametric design could be quantified using various approaches like smallest enclosing hypersphere, convex hull, centroid distance, Determinantal Point Process (DPP) \cite{regenwetter2023beyond,brown2019quantifying}. However, the generated design might be novel and well-spread but may not be realistic. Thus, the design feasibility could be quantified by performing finite element analysis on the generated actuator design \cite{yoo2021integrating}. To the best of author's knowledge, no published study has addressed the importance of generative design in soft actuators and conducted a feasibility test for its design. 

% Once, the data is avalaibale the generative design could lead to similar existing design set. Thus, we need to define novelty and diversity in the generated design. Also, the generated design may be novel but not feasle. Thus, we need to also check the feasibility of the design. As a first approach we try to use FEA to judge the feasibility of the novel and diverse design. Add papers related to {novelyt, diversity, feasibility}. Add papers showcasing FEA of Pnu-net actuators.

In the present study, the pipeline to create novel designs of soft actuators based on generative algorithm is discussed. First, a synthetic dataset of soft Pneu-net actuator is created using data augmentation to train the generative model. Second, Gaussian mixture model has been applied to learn the design parameter space and generate novel designs. Next, the generated designs are visualized in Computer-Aided Design (CAD) software and performance of new actuators are analyzed using Finite Element (FE) software, ABAQUS 2017 (Dassault Systems Simulia). An in-house python script automates the complete process, starting from generating novel design to performing design feasibility test. Further, the algorithm evaluates the novelty, uniformity, spread and feasibility of the generated design and showcased its importance.
% Further, the generative algorithm is compared with the existing designs of soft pneumatic actuators. \textcolor{red}{May need to add few sentences here as well based on performances comparison.}

\section{Generative Design Methodology}
\label{section-2}

\begin{figure}[h]
      \centering
      {\includegraphics [trim=0cm 0cm 0cm 0cm, clip=true,width=1 \linewidth]{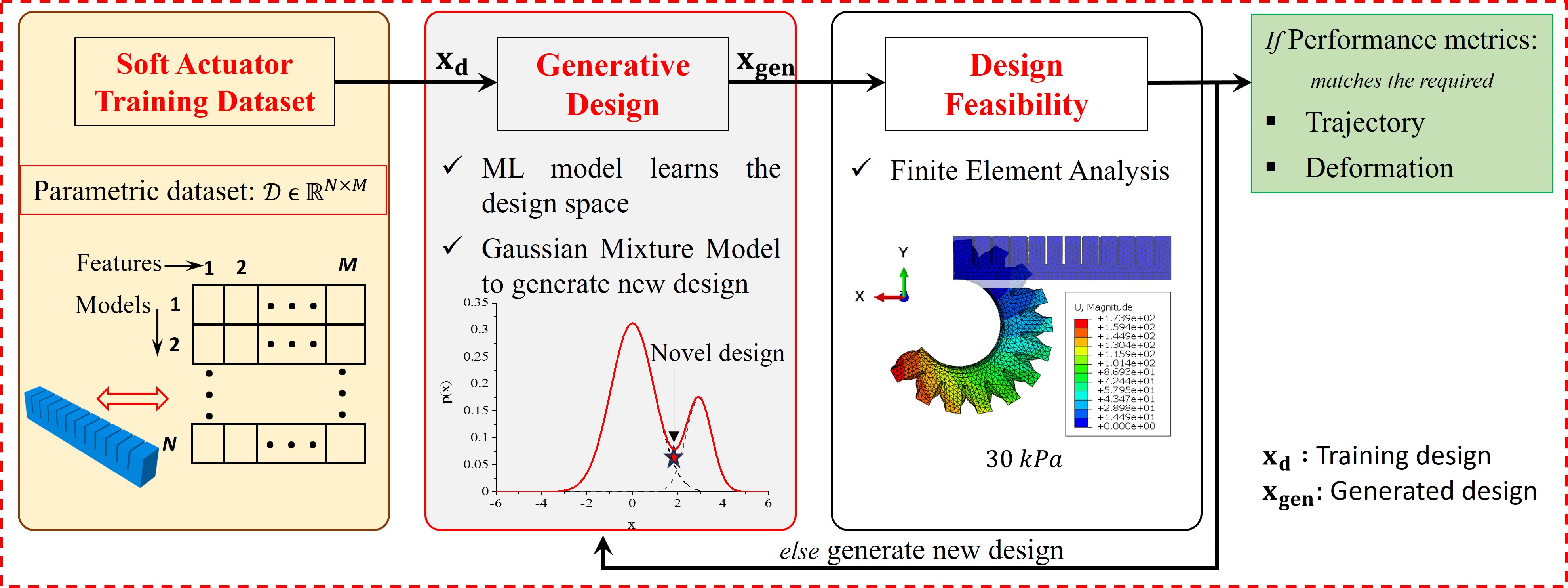}}
      \caption{Flow of Performance-augmented Generative Design (PaGD) methodology for the design of soft actuators. }
      \label{fig:1}
\end{figure}

This section discusses the Performance-augmented Generative Design (PaGD) methodology in detail. Fig. \ref{fig:1} shows the pipeline of PaGD for the design of soft actuators. Firstly, the synthetic data for training the generative model is prepared. Secondly, the generative design section entails creating novel designs using any generative model, like the Gaussian Mixture Model (GMM). The novelty and diversity of the generated design set is evaluated. Lastly, the generated design is passed through a Finite Element (FE) analysis framework to analyze its performance. The performance of the new actuator is assessed and compared to that of the existing actuators. This evaluation determines the feasibility of a new actuator design. The following subsection discusses each block in detail.  

% \textcolor{red}{Should we just say the forward approach? bcz comparing the performance and trying to generate new design will be kind of optimal design problem, which we have highlighted in future work.}
\subsection{Training dataset preparation and preprocessing}
The following subsection addresses the creation of synthetic data and the visualization of soft Pneumatic networks (Pneu-net) actuators through Computer-Aided Design (CAD) software. The unavailability of a design dataset in the literature for soft actuators necessitates the development of a synthetic data set. In this work, the synthetic data is prepared using random sampling and data augmentation. A random sampling technique has been performed over each feature within a specified range. The range of values is chosen based on existing works of Pneu-net actuators \cite{jiang2021modeling}. Data augmentation-based sampling is performed performed by taking into account previous designs \cite{xavier2022soft}. As a result, the design data, $\mathcal{D} \in \mathbb{R}^{N \times M}$, is created where  $N = 11000$ represents models of Pneu-net actuators with $M = 16$ parameters as listed in Table \ref{tab:1}. The design space of the synthetic data is represented as
\begin{equation}\label{eqn:01}
    \mathcal{D} = {\mathbf{\underline{x_d}}}\in \mathbb{R}^{M} \,| {x_d}_i^{(l)} \leq {x_d}_i \leq {x_d}_i^{(u)} \quad \forall \, i=1,2,...M.
\end{equation}
wherein ${x_d}_i^{(l)}$, ${x_d}_i^{(u)}$ are the lower and upper bounds of each parameter ${x_d}_i$. The parameters include geometrical features that are numerical and of categorical type, describing the design of Pneu-net actuators. Moreover, the features may include material properties and different types of cross-sections of the chambers. Herein, two categories, i.e., bending and twisting mode of Pneu-net actuators, are included. The classification is based on the orientation of the chambers. Further, the design data of a soft actuator is used to train the generative model as highlighted in Fig. \ref{fig:1}. 

\begin{table}
\centering
\begin{threeparttable}[b]
\caption{Parameters of Pneu-net actuator used in CAD modeling.}
\centering 
\begin{tabular}{llll}
 	\toprule
&Parameter &Description \\
 \midrule
Independent &$L$  &Length of a rectangular chamber\\
&$W$  &Width of a rectangular chamber\\
&$H$  &Height of a rectangular chamber\\
&$t$  &Wall thickness of a chamber\\
&${t_n}$\tnote{\Cross}  &Distance between two consecutive chambers\\
&$t_h$  &Top head thickness of a chamber\\
&$t_{ab}$  &Wall thickness of in between air chamber\\
&${t_b}$\tnote{\Cross}   &Base layer thickness of the actuator\\
&$N$  &Total number of chambers\\
&$\theta$  &Orientation of the chambers\\
&$\alpha$ &Fraction of helical chambers\\
Dependent &$L_T$  &Total length of the actuator\\
&$N_1$  &Number of helical chambers\\
&$N_2$ &Number of straight chambers\\ 
Category &Mode  &Bending or Twisting modes of actuation\\
&Cross-section &Rectangular \\
\bottomrule
\end{tabular}
     \begin{tablenotes}
       \item \Cross \, \footnotesize In this work, the parameter considered is constant over the length of the actuator. However, it may vary linearly to achieve different geometries.
     \end{tablenotes}
\label{tab:1}
\end{threeparttable}
\end{table}
% \footnotesize{$^{^{*}}$ is considered as constant. However, the parameter can be varied along the length of the actuator.}

\begin{figure}[h]
      \centering
      {\includegraphics [trim=0cm 0cm 0cm 0cm, clip=true,width=0.8\linewidth]{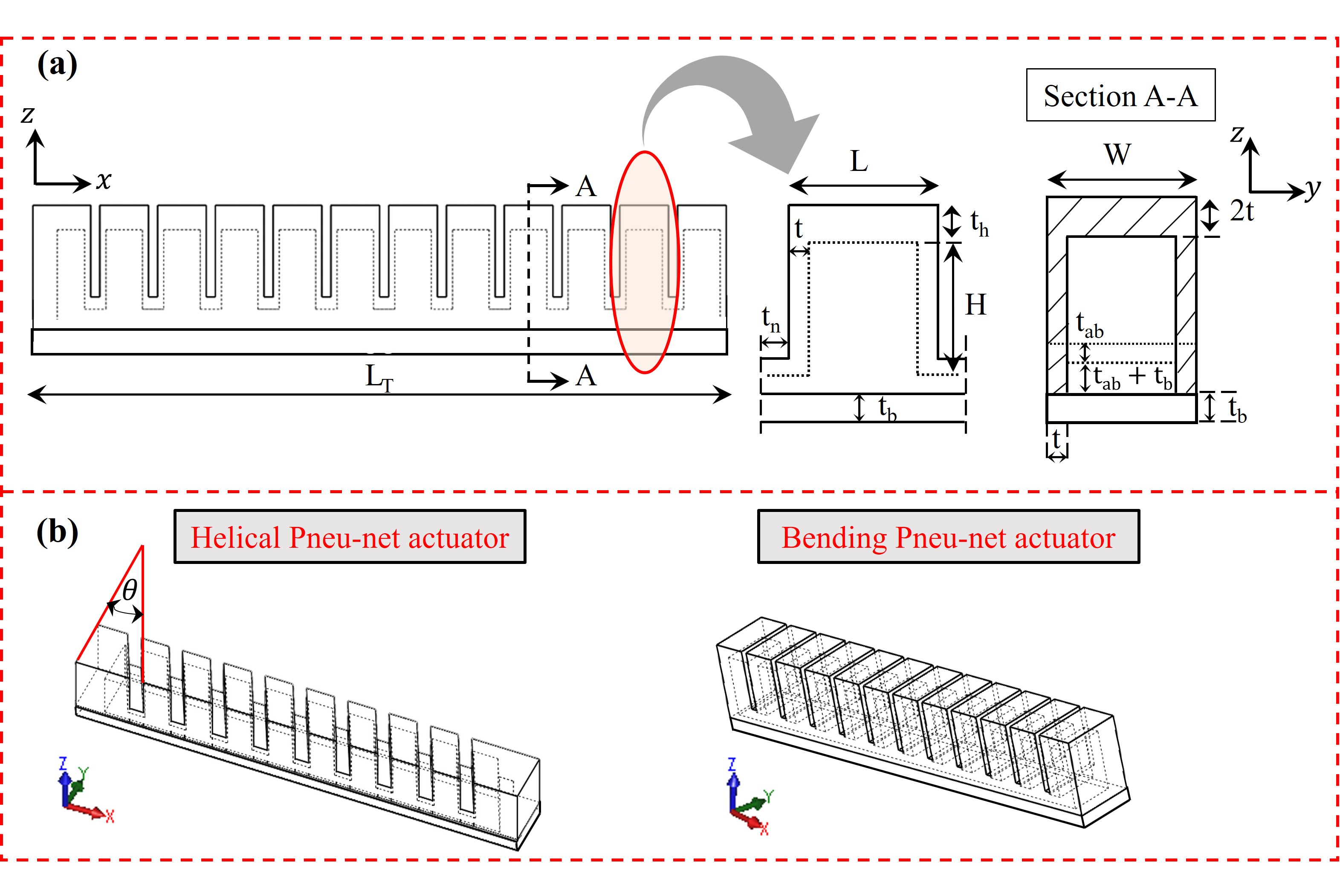}}
      \caption{Visualization of parametric data generated in FreeCAD. Four actuator models representing different geometrical features are visualized in FreeCAD.}
      \label{fig:2}
\end{figure}

Parametric design modeling enables customization of the design using programming languages like Python \cite{machado2019parametric}. One of the significant challenges is to visualize the actuator design for a given parametric dataset. In this work, a parametric design of Pneu-net actuator is visualized using FreeCAD open-source software. An in-house code has been written to automatically generate the CAD files of each model using FreeCAD 0.21.1. This is a necessary step, as CAD files will be a prerequisite for FE simulation at later stages. Fig. \ref{fig:2} (a) shows the design of Pneu-net actuator describing all the geometric parameters/features. The list of parameters are described in Table \ref{tab:1} in detail. Moreover, the two categories of the actuator can be visualised in Fig. \ref{fig:2} (b). It is described in \cite{wang2018programmable} that the orientation of the air chambers can result in twisting from bending mode of actuation.

% shows four randomly printed CAD files from parametric design data of the actuators. Moreover, it is important to highlight that the proposed approach is designed to be applicable to other soft actuators.

Furthermore, for efficient data processing and application of generative models, it is imperative to implement data normalization. The values of certain features of the actuators may exhibit variations in order. The data is shifted to zero mean and unit variances. The dataset comprises diverse data types, including numerical and categorical. Consequently, one-hot encoding is employed on the categorical data type. As a result, one-hot encoding converts categorical into numerical datatype. The final processed parametric design data is further used to train the generative model.

\subsection{Design generation}
This subsection addressed the generation of new designs of soft Pneu-net actuators using machine learning algorithms. Although generative models have been applied to images, design of bicycles, ships, and aerofoils, applying them to the design of soft actuators opens new possibilities. However, the dimensionality of the feature space ($\mathbb{R}^M$) of soft actuators may be large, and that defies straightforward visualization. Also, generative models need to operate on a reduced space rather than a whole feature space. Thus, a dimensional reduction is performed from a high-dimensional feature vector into 2-dimensional space. One approach to achieve this is by applying t-Distributed Stochastic Neighbor Embedding (t-SNE) \cite{van2008visualizing}. t-SNE projects the data onto low-dimensional embedding space, preserving the proximity of similar designs while creating separation between dissimilar ones. The approach uses different similarity measures to construct a similar set of probability distributions. In other words, it represents various clusters present in the data using similarity measures. This may help designers in manually select a point in embedding space which can also represent a novel design. The selected data in latent space needs to be decoded and then Pneu-net actuator can be visualized. Further, the reduced-dimensional feature vector is used to train the generative model.

The novel designs of Pneu-net actuators are generated using the Gaussian Mixture Model (GMM). GMM is a generative model representing the data as a mixture of several Gaussian distributions, providing a flexible framework for capturing complex patterns and generating new data based on learned distributions \cite{bishop2006pattern}. In other words, it models probability distributions by expressing them as a weighted sum of Gaussian components, each defined by mean and covariance parameters. The detail steps involved in GMM algorithm is explained for the reader comprehension. Let's consider the same design data set, $\mathcal{D}$, as described in Eq. \ref{eqn:01}. The probability density function for $K$ Gaussian components is expressed as 
\begin{equation}
    p(\mathbf{\underline{x}} | \Theta) = \sum_{j=1}^{K} w_j \cdot \mathcal{N}(\mathbf{\underline{x}}| \mu_j, \Sigma_j),
\end{equation}
where $w_j$ is the weight associated to $j^{th}$ Gaussian component, $\mu_j$ is the mean vector, and $\Sigma_j$ is the covariance matrix. The probability distribution of M-variate Gaussian is given by
\begin{equation}
    \mathcal{N}(\mathbf{\underline{x}}| \mu_j, \Sigma_j) = \frac{1}{\sqrt{(2\pi)^M |\Sigma_j|}}\exp{\left(-\frac{1}{2}(\mathbf{\underline{x}}-\mu_j)^T\Sigma_j^{-1}(\mathbf{\underline{x}}-\mu_j)\right)}.
\end{equation}
The model parameters \{$w_j, \mu_j, \Sigma_j$\} is collectively expressed as $\Theta$. Now, we introduce a $K$-dimensional binary random variable $z$ which has $z_j$ equal to $1$ and rest elements as zero. The marginal distribution over $z$ specified in
terms of the weights $w_j$ is given by 
\begin{equation}
    p(z_j=1)=w_j, \quad \text{where} \, w_j \in \{0,1\} \, \text{and} \, \sum_{j=1}^{K} w_j=1.
\end{equation} 
Assuming the data points are drawn independently from the distribution, the log likelihood function is defined as
\begin{equation}
    \ln{p(\mathbf{\underline x}|\Theta)} = \sum_{i=1}^{N} \ln \Bigg[\sum_{j=1}^{K} w_j \cdot \mathcal{N}(x_i| \mu_j, \Sigma_j)\Bigg].
\end{equation}
The goal of GMM is to maximize the above log likelihood function and Expectation-Maximization (EM) algorithm is to perform the optimization. In EM, firstly, the model parameters ($\Theta$) is initialized and log likelihood is calculated. Secondly, in E-step , we evaluate the conditional probability of $z$ given $x$,i.e, $\hat{\gamma}(z_{ij})$ is given by 
\begin{equation}
    \hat{\gamma}(z_{ij})=\frac{w_j \cdot \mathcal{N}(x_i| \mu_j, \Sigma_j)}{\sum_{k=1}^{K} w_k \cdot \mathcal{N}(x_i| \mu_k, \Sigma_k)}
\end{equation}
Next, in M-step, the model parameters ($\Theta$) are updated based on $\hat{\gamma}(z_{ij})$. Lastly, the log likelihood is calculated with the updated parameters and the convergence is checked. The convergence is achieved when the difference of last two values of log likelihood function is less than tolerance. it is desired to optimize the model parameters such that the training data is represented well. 

\begin{figure}[h]
      \centering
      {\includegraphics [trim=0cm 0cm 0cm 0cm, clip=true,width=0.8\linewidth]{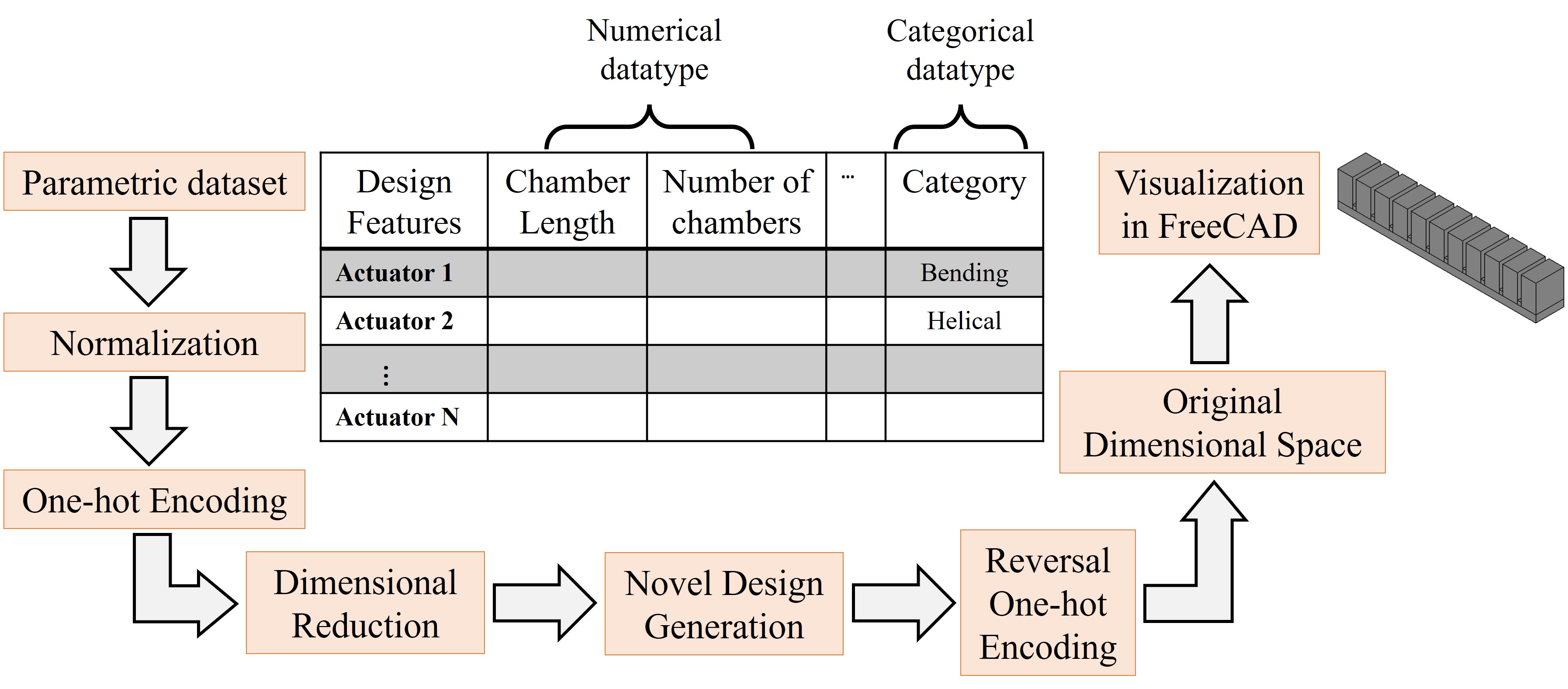}}
      \caption{Flow diagram representing the visualization of a newly generated parametric data in FreeCAD open-source software.}
      \label{fig:3}
\end{figure}

Initially, GMM is trained using existing or synthetic data, capturing diverse distributions for various Pneu-net actuators. Subsequently, new actuator designs are generated by random sampling, featuring bending, twisting, or mixed modes of actuation. One of the bottlenecks of the generative design methodology is the novelty of the generated design data. The assessment of novelty is often subjective. It is easier for the domain expert to judge novelty. However, the presence of domain experts may not be assured in all scenarios. One method to address this issue involves assessing the novelty of the generated data using the similarity measures as discussed in \cite{regenwetter2023beyond}. In the present work, the nearest datapoint approach is used to describe the design novelty. The approach signifies the proximity of a generated sample to its closest datapoint. The numerical value of novelty ($d_{new}$) of the generated design data set is estimated as\\
\begin{equation}
    d_{new} = \frac{1}{|D_{gen}|}\sum_{i=1}^{D_{gen}} \, min [D(\mathbf{\underline x_i}, \mathbf{\underline x_m} )], \quad \mathbf{x_m} \in D_{gen},
\end{equation}
wherein $D(\mathbf{\underline x_i}, \mathbf{\underline x_m} )$ represents Euclidean distance between the generated data from every training data. $D_{gen}$ is the novel generated design data set and $|D_{gen}|$ is the cardinality of the generated set. The uniformity and spread of the generated data set also needs to be quantified. The volume of the convex hull is used as a diversity measure of the generated design data \cite{brown2019quantifying}. The convex hull is often referred to as the two-dimensional shape that an elastic band would form when it is stretched around all the points in a given data set. Further, the newly generated design can be visualized using FreeCAD software. 
% A novel actuator design can be randomly sampled from the GMM distribution. The novel design may choosen by interpolating between different clusters of data. 
Fig. \ref{fig:3} shows the flow of visualization of the generated design data set in FreeCAD. However, the generated design should pass through some performance metric so that its usability can be judged. To accomplish this, we have a design feasibility block, elaborated upon in the following section.

\subsection{Design feasibility}
In this subsection, the performance of a newly generated actuator design is evaluated. A Finite Element (FE) analysis is performed on the generated Pneu-net actuator. Python scripting in ABAQUS 2017 (Dassault Systems Simulia) has been employed to automate the entire analysis procedure. The actuator geometry consists of two parts, namely the top body and the base layer. The top body, which is the main actuator, consists of the air chambers. The base layer is the inextensible layer reinforced with embedded paper in between two layers  \cite{mosadegh2014pneumatic}. The actuator geometry is made of elastomer and discretized using 10-noded tetrahedral elements with a hybrid formulation (C3D10H). The embedded paper is modeled as shell element and discretised using 3-noded triangular shell element (S3). Further, the simulation is performed assuming the left boundary is fixed and applying different pneumatic pressures to the inner walls of the chambers. FE analysis of the generated actuator access its fitness for application. The parameters includes mechanical characterstics of the soft actuators. This comprises of the blocked force, maximum bending angle, strain rate, work density. The performance metrics can eliminate the bad or faulty newly generated design. The method can be extended to evaluate the performance of a soft robot consisting of these actuators. For example, the efficacy of a well-designed soft gripper, composed of Pneu-net actuator, can be evaluated by considering the required degrees of bending, twisting, or extension for each actuator, or the force requirement to hold an object. This open up the possibility of discovering multimodal soft actuators not based on human intuition only but also based on existing design data.

% \begin{algorithm}
% 			\caption{Non-Linear Inversion algorithm} \label{algo:NLI}
% 			\begin{algorithmic}[1]
% 				\vspace{0.1cm}
% 				\Function{Main}{}
% 				\State read data file
% 				\State \textbf{Input:} \texttt{P}, \texttt{NCA}, $\boldsymbol{u}_{m}$, $\omega$, $\rho$
% 				\State \textbf{Output:} $\boldsymbol{\theta}_{opt}$
% 				\State set \texttt{tDoF} = total Degrees of Freedoms (DoF)
% 				\State set $\boldsymbol{\theta}_{0}$ = Initial guess of material parameters
% 				\State set bounds = range for $\boldsymbol{\theta}_{opt}$ \Comment{Optional}
% 				\State set stopping criterion \Comment{\texttt{\#}Iters, \texttt{\#}FunEvals, TolX, TolFun}
% 				\While{stopping criterion}
% 				\State call \Call{Optimization\_module}{\textproc{Objective\_function},bounds}
% 				\State update $\boldsymbol{\theta}{n+1}$ from $\boldsymbol{\theta}{n}$ using Eq. \ref{eq13} \Comment{Using MATLAB function \textit{fminunc}}
% 				\State \Return $\boldsymbol{\theta}_{n+1}$
% 				\EndWhile
% 				\EndFunction
% 			\end{algorithmic} 
% \end{algorithm}
% Based on FE analysis we evaluate its blocked force and max bending angle for a given pressure. 

\section{Results and discussions}\label{section-3}
The current section analyzes the Performance-augmented Generative Design (PaGD) model. The algorithm is applied to perform dimensional reduction and generate novel parametric design data. The novel generated designs are visualized using FreeCAD. Next, the performance of the generated actuator is evaluated using Finite Element (FE) simulations in ABAQUS 2017 (Dassault Systems Simulia).\\

% Figures needed:\\
% \textcolor{red}{Should we show failure design cases in context of Fig. 5? Should we need to give algorithm of full process from generation to FEA?}
\subsection{Evaluating machine learning models}
\begin{figure}[h]
\centering
\begin{subfigure}{.49\textwidth}
  \centering
  {\includegraphics[trim=0.2cm 0.2cm 0.4cm 0.1cm, clip=true,width=.9\linewidth]{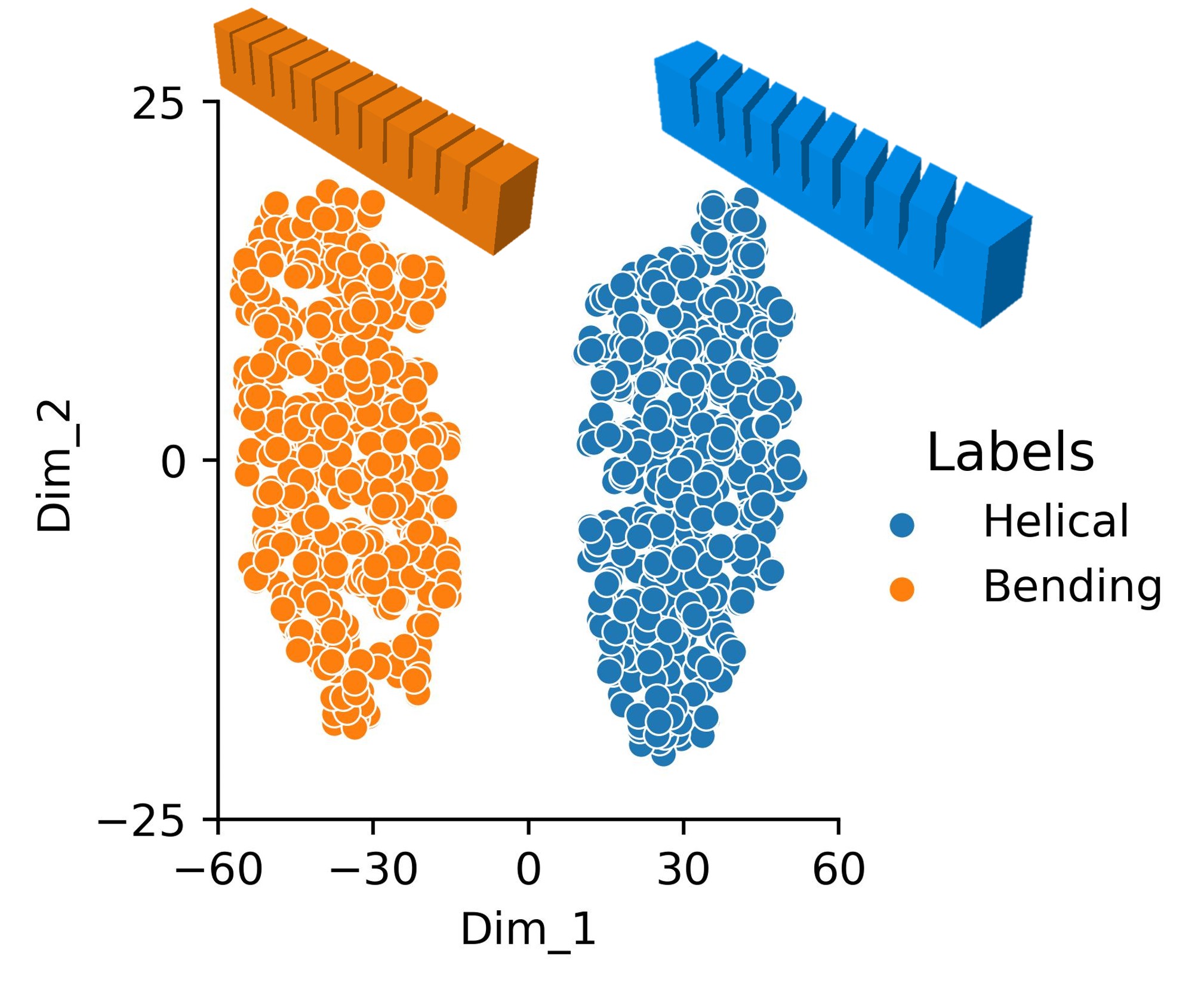}}
   \caption{}
   \label{fig:sub2}
\end{subfigure}
\begin{subfigure}{.49\textwidth}
  \centering
  \raisebox{0.5 cm}{\includegraphics[trim=0cm 0cm 0cm 0cm, clip=true, width=1\linewidth]{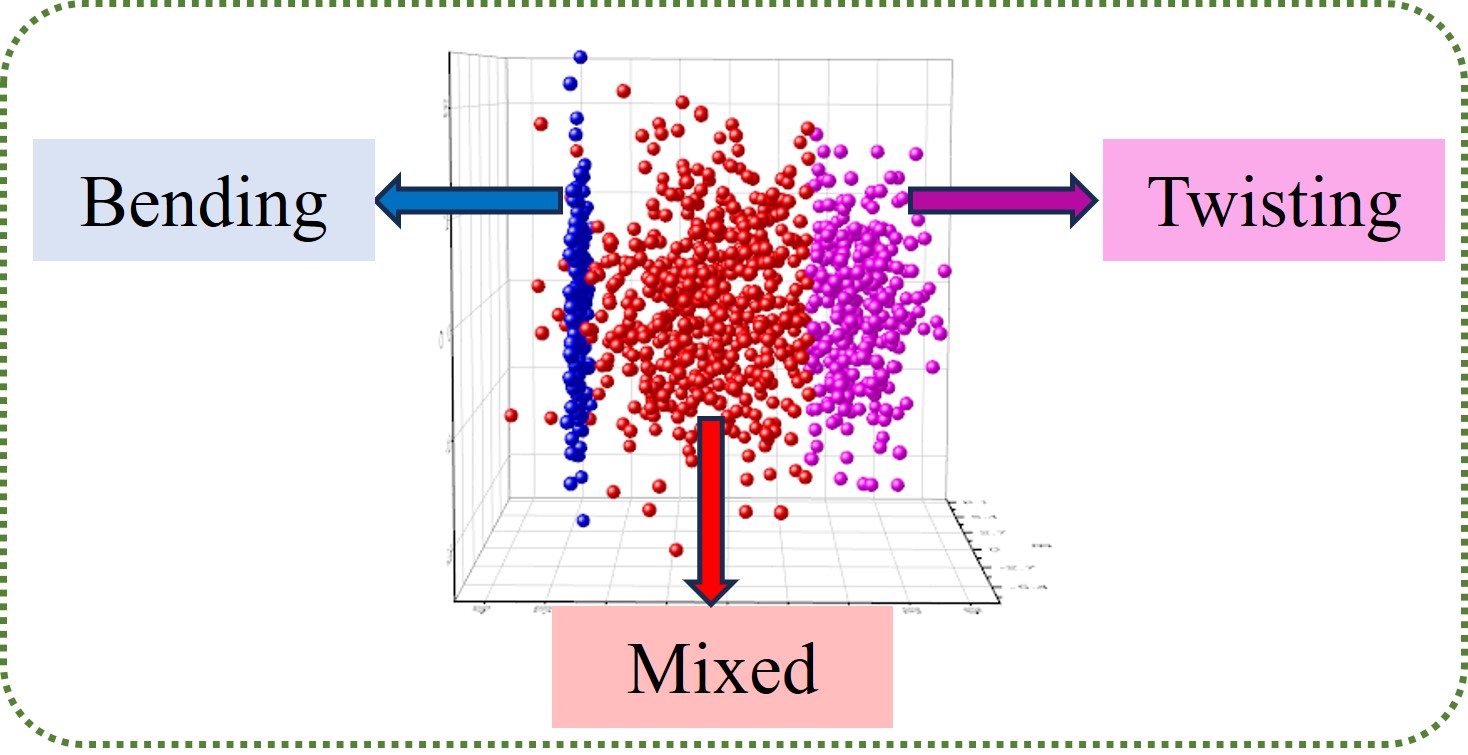}}
  \caption{}
  \label{fig:sub1}
\end{subfigure}
\caption{(a) Visualization of pneu-net actuators using t-Distributed Stochastic Neighbor embedding (t-SNE) of parametric data. Dim\_1 and Dim\_2 are two embedding dimensions. (b) Gaussian mixture model used to learn the distributions of the data. Random sampling generates a novel design from the distribution.}
\label{fig:4}
\end{figure}

Dimensional reduction is generally applied to visualize n-dimensional data into a reduced 2- or 3-dimensional feature space. The dimensional reduction using t-SNE was performed on the synthetic dataset of the Pneu-net actuators. t-SNE is advantageous in visualizing high-dimensional data as it effectively preserves local similarities, offering insights into complex relationships. Its ability to reveal clusters and patterns makes it a valuable tool for exploring intricate data structures in various fields. Fig. \ref{fig:4} (a) represents a randomly sampled thousand data points in the embedded space. t-SNE clearly separates the two different clusters of actuators. The perplexity of the t-SNE model is set at 30. In Fig. \ref{fig:4} (a), Dim\_1 and Dim\_2 represent two embedded spaces. The embedded space values do not have any physical significance. Moreover, the reduced dimension set of data is fed into the generative algorithm for training the model.

Gaussian Mixture Model (GMM) is applied to generate novel designs of Pneu-net actuators with different modes. Fig. \ref{fig:4} (b) represents three clusters of Pneu-net actuator design in reduced dimension, differentiating the modes of the actuator present in the generated design data. Herein, we observed that three Gaussian distributions of the generated design can be obtained with prior information of only two categories of Pneu-net actuator. It highlights that GMM could generate novel data with mixed mode of Pneu-net actuator. The mixed mode of the actuator is a combination of a bending and twisting actuator. The novel design of mixed Pneu-net actuator has the potential to be applied to design bioinspired grippers. The geometry of these actuators enhances the flexibility and maneuverability to the soft gripper design. Consider several possible grasping motions of the human hand, like cylindrical grasping or hooking. It is important to note that both needs different gripper design. The former requires a constant curvature bending over the entire length of the actuator, however, the latter requires bending at the tip of the actuator. These designs of soft grippers can be obtained by altering the stiffness, actuation input, or geometry. The alteration in the geometry by varying the parameter in the design data could result in generating a novel soft gripper for desired application, as discussed in Table \ref{tab:1}. For example, the variability in the distance between two chambers or the thickness of the base material could result in different grippers. Similarly, the variability of the material properties in the parameter data set could be included to obtain the variable stiffness gripper. Moreover, the increase in contact area with more complex configurations, like the mixed Pneu-net actuator could enhance the effectiveness of the grasping capabilities. This emphasizes the significance of parameterized design data and utilizing generative algorithms to generate new designs of soft actuators.

\begin{figure}[h]
      \centering
      {\includegraphics [trim=0cm 0cm 0cm 0cm, clip=true,width=0.6\linewidth]{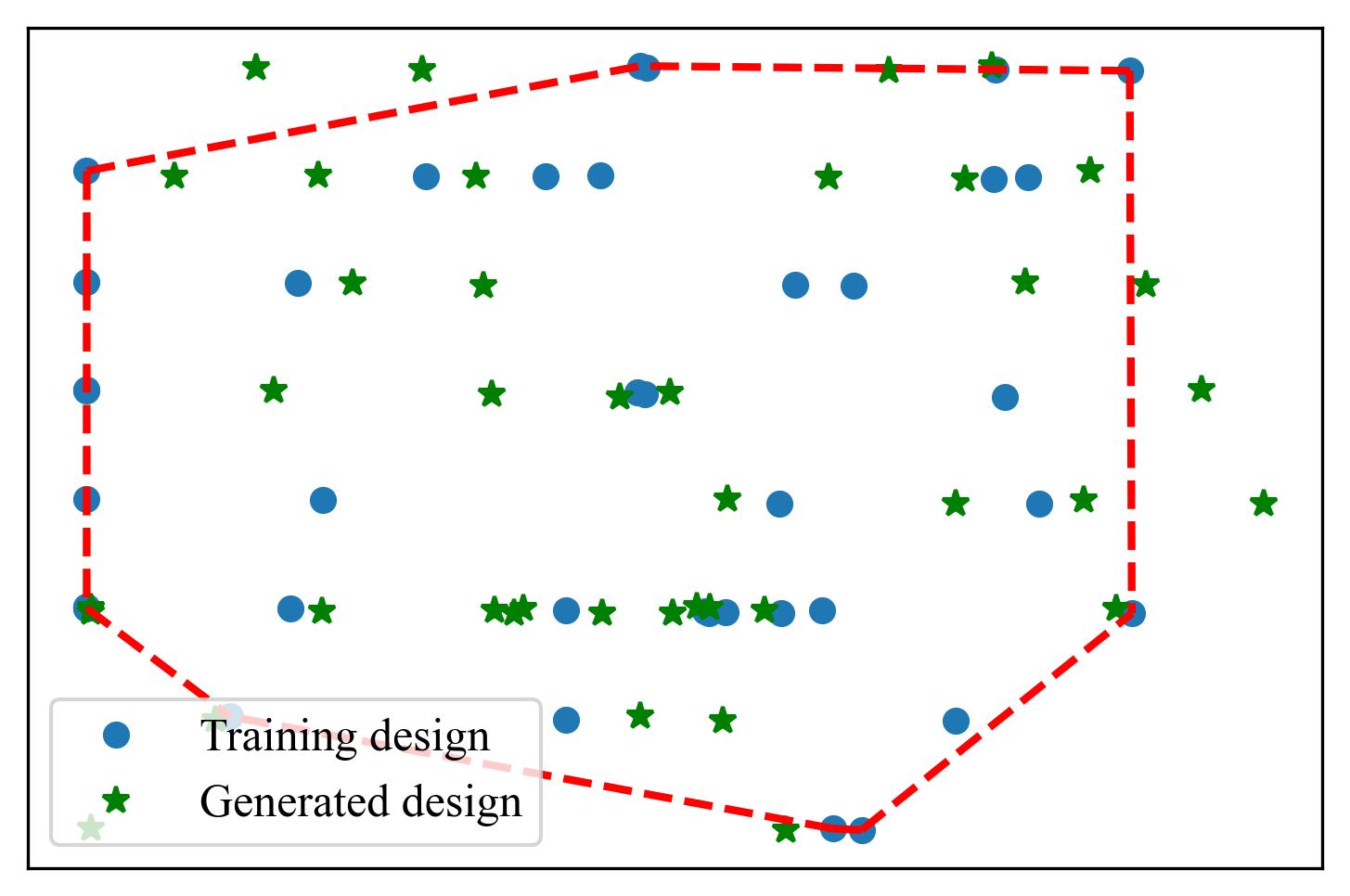}}
      \caption{A convex hull of training design data set. Comparison showing the diversity of the generated design data set with respect to training data.}
      \label{fig:51}
\end{figure}

Further, a randomly sampled thousand novel designs are generated using the trained GMM. The generated design can be visualized using the flow described in Fig. \ref{fig:3}. It is essential to understand that the designs generated are within the distribution. However, the novelty of the generated data can be estimated using the nearest datapoint method, as discussed in subsection 2.2. In our case, the novelty score ($d_{new})$ for an average of 1000 randomly generated data sets is 0.014. Essentially, this indicates the distance of the generated data point from the training data; a higher value is favourable in this context. Further, the diversity of the generated design set is measured by volume of convex hull. Fig. \ref{fig:51} shows the convex hull of randomly sampled $50$ training and generated design data. The area enclosed (2-D case) by the hull represents the qualitative understanding of spread and uniformity of the data. The larger area enclosed ensures a better spread of the data set. However, one can ensure the uniformity of the data set by visual assessment of the convex hull. The resulting parametric design data may possess novelty and diversity, but they do not exhibit high quality performance. Hence, it is essential to perform a simulation over the newly generated design, as discussed in next subsection.

% \begin{figure}[h]
%       \centering
%       {\includegraphics [trim=0cm 0cm 0cm 0cm, clip=true,width=0.7\linewidth]{Fig10.jpg}}
%       \caption{An illustrative example showing various human hand functionality. A motivation for multimodal Pneu-net actuator and need for automated design of universal soft gripper.}
%       \label{fig:52}
% \end{figure}

% \begin{figure}[h]
%       \centering
%       {\includegraphics [trim=0.2cm 0.2cm 0.4cm 0.1cm, clip=true,width=0.65\linewidth]{Fig21.jpg}}
%       \caption{Visualization of pneu-net actuators using t-Distributed Stochastic Neighbor embedding (t-SNE) of parametric data. Dim\_1 and Dim\_2 are two embedding dimensions.}
%       \label{fig:4}
% \end{figure}

% \textcolor{red}{How would we generate multimodal actuator from single mode using GenAI? Can we do that? Done\\
% Compare the performance (blocked force, bending angle) of single mode, generated mixed mode with exiting work.}
\subsection{Performance of generated design}
\begin{figure}[H]
\centering
\begin{subfigure}{.6\textwidth}
  \centering
  {\includegraphics[trim=0cm 0.2cm 0cm 0cm, clip=true,width=1.1\linewidth]{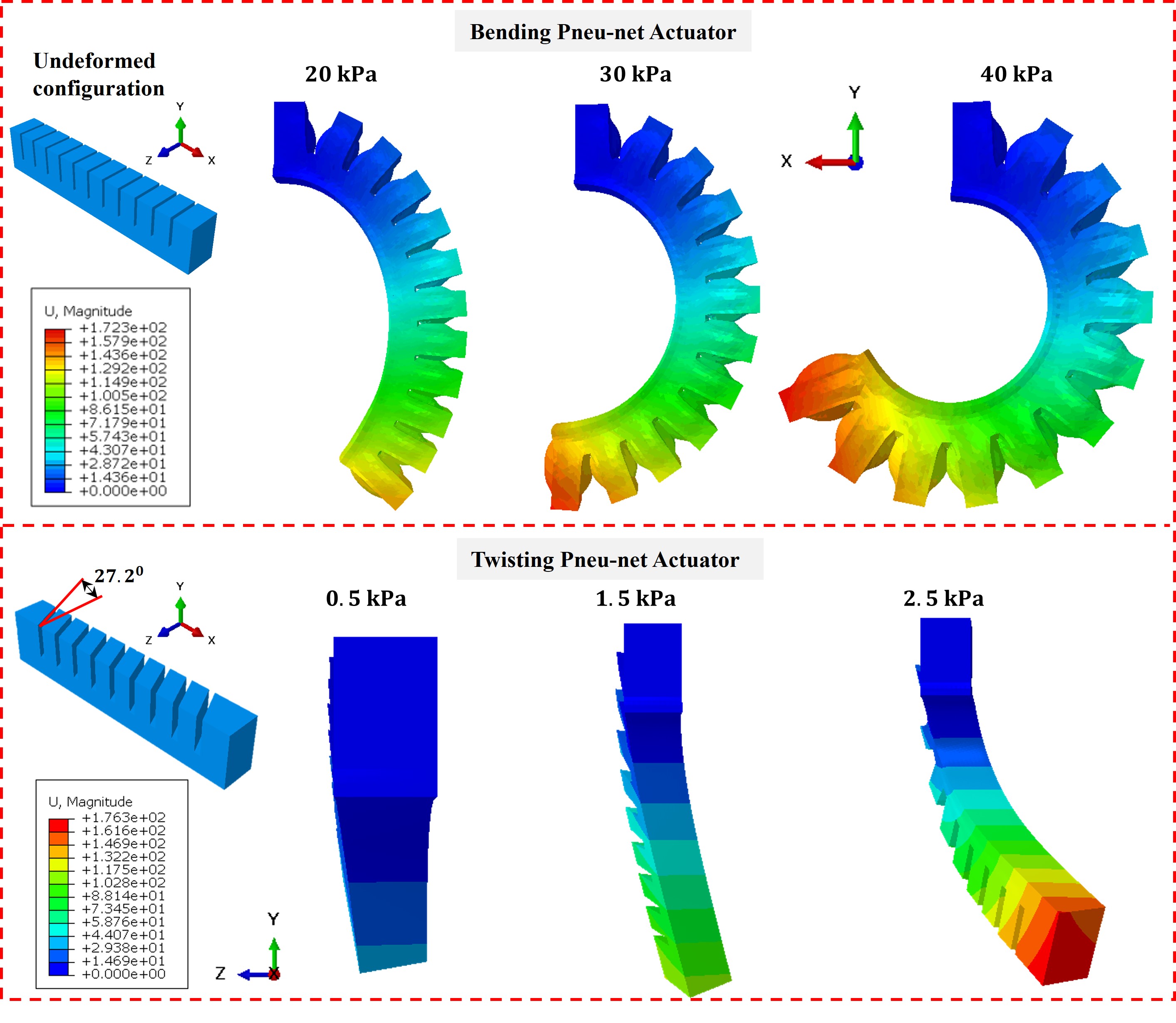}}
   \caption{}
   \label{fig:sub2}
\end{subfigure}
\begin{subfigure}{.6\textwidth}
  \centering
  {\includegraphics[trim=0cm 0cm 0cm 0cm, clip=true, width=1.1\linewidth]{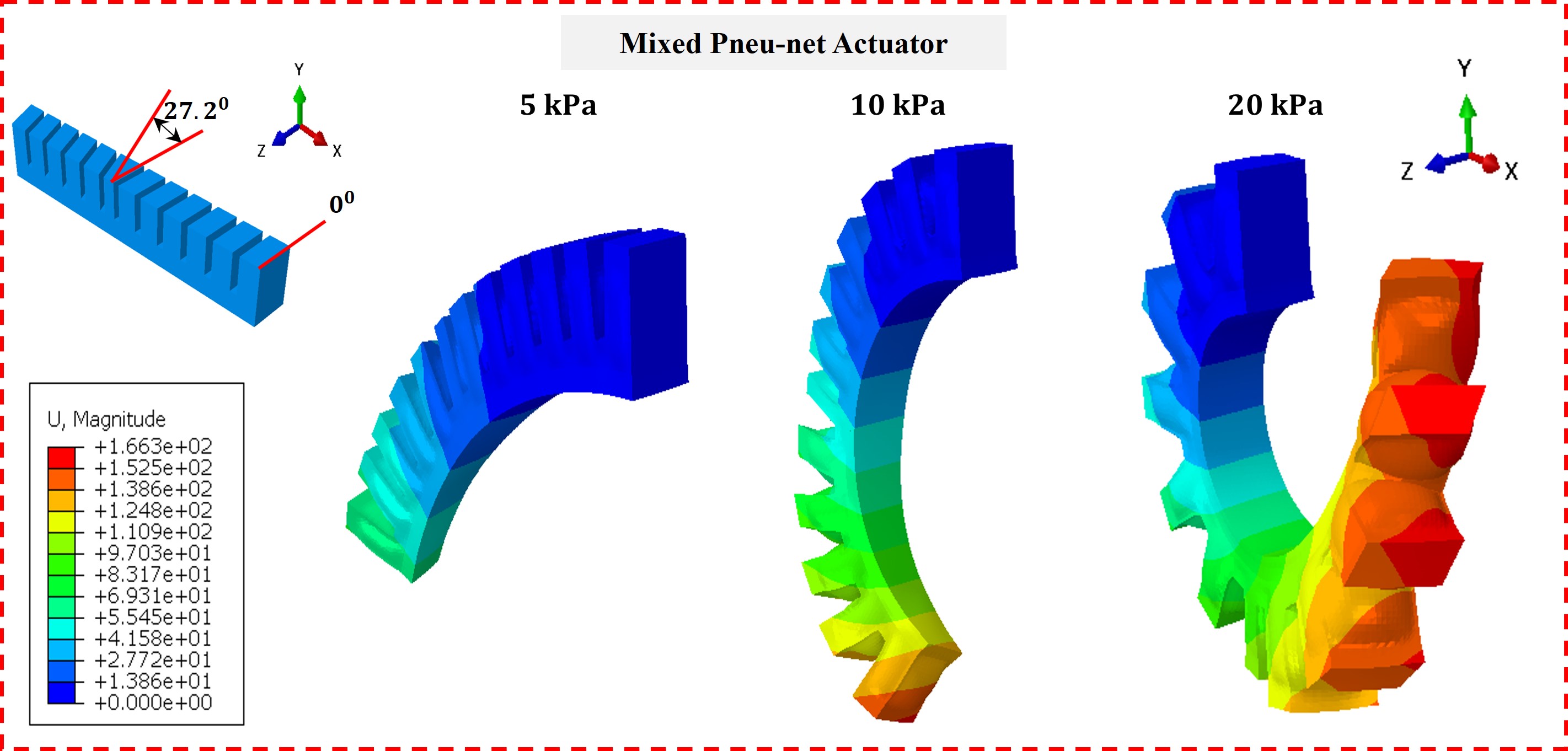}}
  \caption{}
  \label{fig:sub1}
\end{subfigure}
\caption{Finite Element simulation of one of the newly generated (a) bending and twisting actuator, (b) multimodal Pneu-net actuator using ABAQUS Python scripting.}
\label{fig:5}
\end{figure}

The novel generated designs of Pneu-net actuators are simulated using commercially available FE software, ABAQUS 2017. The geometric dimensions of the actuators used in the simulation is described in Table \ref{tab:2}. The complete simulation process is automated using a Python script. The generated design alters the parameters, resulting in a new configuration of the Pneu-net actuator with distinct geometric features. As a result, some configurations might fail to actuate under the specified loading conditions, and others may exhibit non-convergence. Henceforth, it is imperative to subject the generated design to a simulation verification process. Fig. \ref{fig:5} shows a simulation of three novel generated design from GMM. One of the configurations shows a bending motion, the other exhibits a twisting motion, and third one exhibits an in-plane bending and out-of-plane twisting motion. It is observed that the deformation increases with an increase in pressure, as expected. Fig. \ref{fig:5} (b) shows the deformation of multimodal Pneu-net actuators which can be generated by combining straight and inclined chambers. The generated design parameters include a combination of straight and inclined chambers that creates a combined bending and twisting motion when pressurized air is applied. This highlights the utility of generative design in context of soft actuators. PaGD is versatile and opens up opportunities for application in the design of other soft actuators.

\begin{figure}[H]
      \centering
      {\includegraphics [trim=0cm 0cm 0cm 0cm, clip=true,width=1\linewidth]{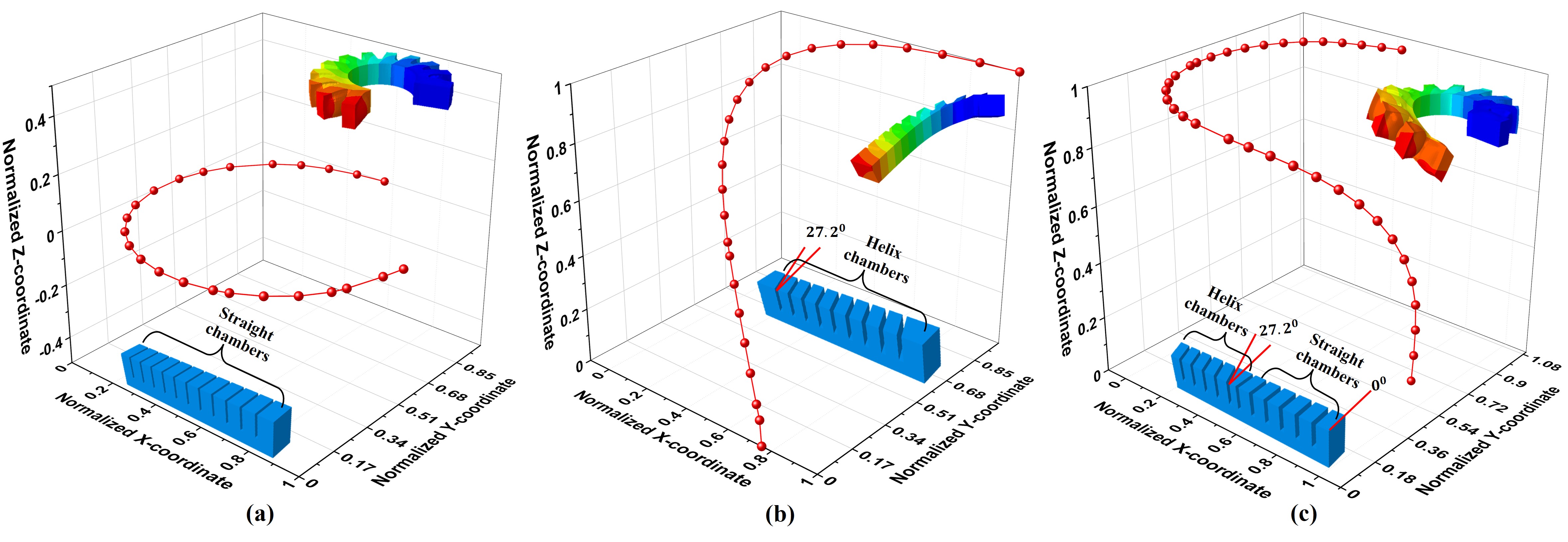}}
      \caption{Trajectory of the deformed shape of newly generated Pneu-net actuator showcasing different modes of actuation. (a) The straight chambers shows purely bending mode, (b) The helical oriented chambers shows in-plane bending and out of plane twist, and (c) The combined Pneu-net actuator shows an effective bending as well as twisting mode (multi-modal) actuation on application of pneumatic pressure.}
      \label{fig:7}
\end{figure}

Furthermore, the performance metrics of the actuator should no longer depend solely on a single scalar metric to assess the design \cite{regenwetter2023beyond}. Occasionally, it is desirable to have a qualitative understanding of the actuator's design which could be obtained by high fidelity simulation. For an example, considering  a soft gripper made of Pneu-net actuators needs a certain geometric profile to hold the object firmly \cite{jiang2021modeling}, as discussed in previous subsection 3.1. Fig. \ref{fig:7} shows the trajectory of the mixed Pneu-net actuator that could be beneficial in gripping rather than having just one form of chambers. Hence, the FE simulations of the generated design provides a leverage over the existing automated design approach, which uses a scalar quality metric. 

\begin{table}[h]
\caption{Geometric dimension of the generated design used for simulation.}
\centering 
\begin{tabular}{llll}
 	\toprule
Parameter &Bending Pneu-net &Twisting Pneu-net  &Mixed Pneu-net \\
 \midrule
$L$ (mm)  &9.51 &7.83 &8.01\\
$W$ (mm)  &15.2 &16.55 &15.12\\
$H$ (mm)  &13.01 &8.5 &12.98\\
$t$ (mm)  &4.02 &0.76 &1.49\\
$t_n$ (mm)  &1.5 &3.89 &2.8\\
$t_h$ (mm)  &3.95 &3.05 &4.07\\
$t_{ab}$ (mm)  &1.95 &1.89 &2.05\\
$t_b$ (mm)  &2.12 &2.4 &1.97\\
$N$  &12 &8 &12\\
$\theta$ (degree)  &0 &27.2 &27.2\\
$\alpha$ &0 &1 &0.5\\
\bottomrule
\end{tabular}
\label{tab:2}
\end{table}

% \raisebox{1.75 cm}
% \begin{figure}[h]
%       \centering
%       {\includegraphics [trim=0cm 0cm 0cm 0cm, clip=true,width=0.8\linewidth]{Fig6.jpg}}
%       \caption{Finite Element simulation of one of the newly generated Pneu-net actuator using ABAQUS Python scripting.}
%       \label{fig:5}
% \end{figure}

% \begin{figure}[H]
%       \centering
%       {\includegraphics [trim=0cm 0cm 0cm 0cm, clip=true,width=0.8\linewidth]{Fig7.jpg}}
%       \caption{Finite Element simulation of one of the newly generated multimodal Pneu-net actuator using ABAQUS Python scripting.}
%       \label{fig:6}
% \end{figure}

\subsection{Limitations and Future Directions}
% \textcolor{red}{Should we include Padgan paper and MO-Padgan paper? Will it reduce the value of our work?\\}
A significant limitation lies in the potential bias of the novel design towards the existing distribution, given the minimal dataset employed in the current study. Our future objectives encompass expanding the dataset by including new features in data set and making it publicly accessible. This aligns with principles of scientific transparency and reproducible contribution to the broader research community by providing a comprehensive and diverse resource for soft actuator design exploration. Although, the PaGD model have been  applied to design soft pneumatic actuators, it is important to note that the concept is applicable to other types of soft actuators design as well. Moreover, it is possible that the novel actuator design is out of distribution but fails to generate the desired performance. The high fidelity simulation of the generated design is needed but is computational expensive. An alternative approach towards reduced order modeling of the novel design needs to be explored. Further, it is seen that optimal design of soft actuators are task specific. For example, a Pneu-net actuator optimally designed to perform gripping may not be suitable or perform best for locomotion robots. 
% This clearly differentiates the approach of optimal design from generative design. Fig. \ref{fig:8} shows that the feasible design set includes various possible optimal design for different task. A user may choose the optimal novel design by performing optimization for a specific task. Hence, a multi-objective constrained optimization can be coupled with the generative design approach and will be included in our future study. 
  
% \textcolor{red}{Talk about the limitation of high fidelity simulation like FEA, talk about PadGAN paper. Multi-objective Optimization work. Refer to the figure PaGD+Optimization.Discuss the perspective.}

% \begin{figure}[H]
%       \centering
%       {\includegraphics [trim=0cm 0cm 0cm 0cm, clip=true,width=0.65\linewidth]{Fig81.jpg}}
%       \caption{Optimal generative design of soft actuators.}
%       \label{fig:8}
% \end{figure}

\section{Conclusion}\label{section-4}
The present work describes the pipeline of an automated performance-augmented generative design of soft pneumatic actuators. A synthetic dataset of Pneu-net soft actuators has been created using data augmentation due to the unavailability of public data. Herein, the parametric design data of Pneu-net actuators is used to train the generative model. However, the approach is general and can accommodate any soft actuator design. The generative algorithm, i.e., the Gaussian mixture model (GMM), is used to generate the novel design of the Pneu-net actuator. An in-house code was written to automatically generate CAD files of Pneu-net actuators from parametric data for visualization, and it was used later in performance analysis. GMM learns the distribution of the design parameter space of two types of Pneu-net actuators, namely, bending and helical type, and generates a mixed type of Pneu-net actuator performing multimodal actuation. Moreover, the convex hull of the generated design data set shows uniformity and spread. Later, a finite element simulation is performed on the newly generated design to predict its response to an applied pressure. The simulation points out that the multimodal actuation could generate a more flexible design of Pneu-net actuators, which could improve soft grippers' manoeuvrability. The multimodal soft grippers could improve the functionality of the existing gripper design. However, the forward approach of learning the parametric design space distribution might or might not provide us with an optimal design for a particular task. Nonetheless,  the optimal novel design for a particular task would be a subset of the generated design data set. Hence, the optimal design integrated with the generative algorithm is our future direction of study. 

% \begin{figure}[H]
% \centering
% \begin{subfigure}{.49\textwidth}
%   \centering
%   \frame{\includegraphics[trim=0cm 0cm 1cm 0.5cm, clip=true,width=.75\linewidth]{BG40.eps}}
% %   \caption{A subfigure}
% %   \label{fig:sub2}
% \end{subfigure}
% \begin{subfigure}{.49\textwidth}
%   \centering
%   {\includegraphics[trim=0.2cm 0cm 1cm 0.5cm, clip=true, width=.75\linewidth]{FigF5}}
%   %\caption{A subfigure}
%   %\label{fig:sub1}
% \end{subfigure}
% \begin{subfigure}{.49\textwidth}
%   \centering
%   {\includegraphics[trim=0cm 0cm 1cm 0.5cm, clip=true,width=.75\linewidth]{FigF9}}
% %   \caption{A subfigure}
% %   \label{fig:sub2}
% \end{subfigure}
% \begin{subfigure}{.49\textwidth}
%   \centering
%   {\includegraphics[trim=0.2cm 0cm 1cm 0.5cm, clip=true,width=.75\linewidth]{FigF6}}
% %   \caption{A subfigure}
% %   \label{fig:sub2}
% \end{subfigure}
% \begin{subfigure}{.49\textwidth}
%   \centering
%   {\includegraphics[trim=0cm 0cm 1cm 0.5cm, clip=true,width=.75\linewidth]{FigF10}}
% %   \caption{A subfigure}
% %   \label{fig:sub2}
% \end{subfigure}
% \begin{subfigure}{.49\textwidth}
%   \centering
%   {\includegraphics[trim=0cm 0cm 1cm 0.5cm, clip=true,width=.75\linewidth]{FigF7}}
% %   \caption{A subfigure}
% %   \label{fig:sub2}
% \end{subfigure}
% \caption{Effect of anisotropy factor on FTCP actuator at various applied step voltage for different fiber angle.}
% \label{fig:9}
% \end{figure}

\section*{Acknowledgements}
The authors greatly acknowledge the computing facilities provided by the Indian Institute of Technology Delhi, India.

\section*{Declaration of competing interest}
The author(s) declared no potential conflicts of interest with
respect to the research, authorship, and/or publication of this
article.

\subsection*{Code Availability}
All the source codes to reproduce the results in this study will be made available to the public on GitHub upon acceptance.

% \bibliographystyle{unsrtnat}
% \bibliography{references}  %%% Uncomment this line and comment out the ``thebibliography'' section below to use the external .bib file (using bibtex) .


\end{document}